\documentclass[conference]{IEEEtran}
\IEEEoverridecommandlockouts
\usepackage{cite}
\usepackage{amsmath,amssymb,amsfonts}
\usepackage{algorithm}
\usepackage{algorithmic}
\usepackage{graphicx}
\usepackage{textcomp}
\usepackage{xcolor}
\def\BibTeX{{\rm B\kern-.05em{\sc i\kern-.025em b}\kern-.08em
    T\kern-.1667em\lower.7ex\hbox{E}\kern-.125emX}}
\usepackage{booktabs}
\usepackage{threeparttable}
\usepackage{graphicx}
\usepackage{array}
\usepackage{multirow}

\begin{document}

\title{Evaluating Temporal Plasticity in Foundation Time Series Models for Incremental Fine-tuning
}

\author{\IEEEauthorblockN{1\textsuperscript{st} Jia Liu}
\IEEEauthorblockA{\textit{School of Software Engineering} \\
\textit{Sichuan University)}\\
Chengdu, China \\
jia\_liu@stu.scu.edu.cn}
\and
\IEEEauthorblockN{2\textsuperscript{nd} Cheng Jinguo}
\IEEEauthorblockA{\textit{National Key Laboratory of Fundamental Science of Synthetic Vision} \\
\textit{Sichuan University)}\\
Chengdu, China \\
2023226040005@stu.scu.edu.cn}
\and
\IEEEauthorblockN{3\textsuperscript{rd} Xia Fang}
\IEEEauthorblockA{\textit{School of Mechanical Engineering} \\
\textit{Sichuan University)}\\
Chengdu, China \\
fangxia@scu.edu.cn}
\and
\IEEEauthorblockN{4\textsuperscript{th} Zhenyuan Ma}
\IEEEauthorblockA{\textit{China Electric Power Research Institute} \\
Beijing, China \\
mazhenyuan@epri.sgcc.com.cn}
\and
\IEEEauthorblockN{5\textsuperscript{th} Yuankai Wu*\footnote{*Corresponding Author}}
\IEEEauthorblockA{\textit{School of Computer Science} \\
\textit{Sichuan University}\\
Chengdu, China \\
wuyk0@scu.edu.cn}
}

\maketitle

\begin{abstract}
Time series foundation models excel at diverse time series forecasting tasks, but their capacity for continuous improvement through incremental learning remains unexplored. We present the first comprehensive study investigating these models' temporal plasticity—their ability to progressively enhance performance through continual learning while maintaining existing capabilities. Through experiments on real-world datasets exhibiting distribution shifts, we evaluate both conventional deep learning models and foundation models using a novel continual learning framework. Our findings reveal that while traditional models struggle with performance deterioration during incremental fine-tuning, foundation models like Time-MoE and Chronos demonstrate sustained improvement in predictive accuracy. This suggests that optimizing foundation model fine-tuning strategies may be more valuable than developing domain-specific small models. Our research introduces new evaluation methodologies and insights for developing foundation time series models with robust continuous learning capabilities.
\end{abstract}


\section{Introduction}

Time series data is a fundamental modality that underpins dynamic systems and plays a crucial role across a wide range of real-world applications, from finance\cite{sezer2020financial} and healthcare\cite{tonekaboni2020unsupervised} to transportation\cite{li2018diffusion} and environmental monitoring\cite{bi2023accurate}. In various domains, the collection of time series data involves diverse types, each characterized by distinct time-varying temporal structures, inter-series correlations, and complex distributions. These unique properties have traditionally driven the development of domain-specific deep learning architectures, each tailored to address the specific challenges of different time series analysis tasks. For example, some studies argue that Temporal Convolutional Networks (TCN) are particularly effective for capturing long-range dependencies\cite{wan2019multivariate,hewage2020temporal,bai2018empirical}, while others emphasize the advantages of Transformer-based models and their variants in modeling intricate temporal relationships and interdependencies across multiple series\cite{zhang2023crossformer,liu2023itransformer}. This diversity of approaches reflects the inherent complexity of time series analysis, where the choice of model often depends heavily on the specific task at hand. Consequently, this task-dependent approach raises the question of whether more generalized, flexible models can be developed to better capture the variety of temporal patterns and correlations present across different domains.

Recently, research in the time series domain has revealed a phenomenon similar to that observed in the fields of computer vision and natural language processing: models exhibit scaling laws\cite{edwards2024scaling}. These scaling laws suggest that as model size and the amount of training data increase, performance continues to improve, often in a predictable, power-law manner. As time series pre-training datasets are no longer a major bottleneck\cite{woo2024unified}, numerous industrial organizations, leveraging their computational power, have developed highly capable foundation time series models and made their parameters publicly available for industry use. These models, such as Time-MoE\cite{shi2024time}, TimesFM\cite{das2023decoder}, and Chnoros\cite{ansari2024chronos}, have demonstrated significantly stronger zero-shot and few-shot capabilities compared to smaller models. 

When foundation time series models are evaluated, the primary focus is often on their zero-shot capabilities\cite{gruver2024large}, where models are tested on entirely new and unfamiliar time series data. Additionally, attention is given to full-shot performance, where the time series data is divided into training, validation, and test sets, with model performance assessed on the test set. However, these evaluation methods are typically conducted in controlled, laboratory-like settings and do not fully capture the complexities of real-world industrial applications.

In actual scenarios, time series data is collected in a dynamic, continuous manner. As time progresses, the data often undergoes significant distribution shifts, driven by changes in the system or external factors\cite{zhou2022time}. For example, in a predictive maintenance scenario, sensor data may remain consistent over time, but a major equipment upgrade or environmental change—such as a shift in production conditions or an unexpected change in machine behavior—can lead to new patterns emerging in the data. Such shifts, which are common in industrial environments, can render models trained on past data less effective when faced with new data that diverges from the historical distribution\cite{wu2024adaptive}. 

Therefore, there is a pressing need for foundation models with continual learning capabilities that can be fine-tuned or updated with newly collected data over time. Such models should be able to adapt to the evolving nature of time series data, accommodating distribution shifts and maintaining robust performance through ongoing learning from the latest data. Unfortunately, as highlighted in a recent work\cite{dohare2024loss}, many deep learning models tend to gradually lose their adaptability in continuous learning environments. This loss of adaptability, or "plasticity" refers to the model's diminishing ability to effectively learn from new data while retaining its previous knowledge. In other words, as these models are incrementally updated with new information, they struggle to generalize to new patterns and may even forget previously learned tasks, a phenomenon known as catastrophic forgetting\cite{mccloskey1989catastrophic}. If time series foundation models were to experience this issue, it would be catastrophic for practical applications. These models, which already require significant computational resources for training, would become inefficient and unreliable over time if they lost the ability to adapt to new data while preserving their performance on previous tasks. This would be a major concern for users, as the ongoing need for resource-intensive retraining could make such models impractical for real-world use, where efficiency and scalability are key requirements.

To evaluate the incremental fine-tuning capability of time series foundation models on new data over time, we have designed a pipeline specifically for time series continual learning. This pipeline allows us to assess the model’s ability to adapt to new temporal observations through continuous fine-tuning. Additionally, we have developed a set of evaluation metrics to comprehensively test the ‘temporal plasticity’ of both mainstream smaller deep models and foundation models, providing a more complete understanding of their performance over time. Our findings reveal that traditional deep learning models suffer significantly from a loss of plasticity, struggling to retain their ability to adapt to new observed data without forgetting prior knowledge. Moreover, as incremental fine-tuning on new temporal observations continues, their performance deteriorates progressively. In contrast, foundation models, such as Time-MoE, trained with large parameters and extensive pre-training data, effectively mitigate this issue. The contribution of this paper is summarized as the following:

\begin{itemize}
    \item To the best of our knowledge, this is the first work to investigate the "temporal plasticity" of time series foundation models using continuous fine-tuning.
    \item Our research also suggests that refining the fine-tuning mechanisms of time series foundation models could prove more valuable than continuing to develop small models tailored to specific domains.
    \item Our experiments provide new options for evaluating the capabilities of foundation time series models, contributing valuable insights to the ongoing development of more robust models for practical use.
\end{itemize}

\section{Related Work}
\subsection{Time Series Foundation Model}

In recent years, foundation models have revolutionized time series analysis by achieving remarkable performance in zero-shot and few-shot tasks through large-scale pre-training on diverse datasets \cite{ye2024survey}. Notable examples include Time-MoE \cite{shi2024time}, which employs a mixture-of-experts mechanism for efficient and scalable temporal modeling; Chronos \cite{ansari2024chronos}, which utilizes tokenization with scaling and quantization to capture periodic and irregular patterns through advanced temporal encoding; Moirai \cite{woo2024unified}, which integrates adaptive fine-tuning to handle dynamic temporal distributions; and models such as GPT4TS\cite{GPT4TS}, Time-LLM\cite{timellm}, and AutoTimes\cite{liu2024autotimes}, which adapt LLMs for time series tasks using cross-attention mechanisms, temporal-text alignment, and semantic prompts to address challenges in zero-shot learning, knowledge transfer, and reasoning. Despite these advancements, current models primarily focus on static evaluation scenarios and often overlook adaptability to dynamic, real-world data. As time series data frequently undergoes distributional shifts due to environmental changes or system upgrades, existing models may lack the ability to continuously fine-tune on newly observed data. 

\subsection{Loss of Plasiticity and Catastrophic Forgetting in Deep Learning}
Deep learning models face two major challenges in continuous training environments: loss of plasticity and catastrophic forgetting\cite{wang2024comprehensive}. The former refers to the model's diminishing ability to learn new information due to over-specialization on recent tasks, resulting in reduced representational diversity and difficulty adapting to new data distributions\cite{mccloskey1989catastrophic}. The latter describes the gradual loss of knowledge from previous tasks as new tasks overwrite earlier representations during gradient updates. Studies have shown that catastrophic forgetting and loss of plasticity interact in continuous learning settings, collectively limiting model performance and adaptability\cite{dohare2024loss}. To address loss of plasticity, Sutton et al. \cite{dohare2024loss} proposed the Continuous Backprop method, which maintains adaptability through stochasticity injection. Techniques such as ReDo\cite{sokar2023dormant} and L2 Init\cite{kumar2024maintaining} further mitigate this issue by reactivating dormant neurons and applying regularization to model parameters. To reduce catastrophic forgetting, approaches like parameter isolation\cite{serra2018overcoming}, knowledge distillation\cite{rebuffi2017icarl}, and experience replay\cite{lopez2017gradient} are widely adopted. Time series models inherently involve dynamic and continuous data collection, often accompanied by significant domain distribution shifts in real-world applications\cite{zhou2022time}.

\section{Problem Formulation}

\subsection{Traditional Training and Inference Pipeline}
Traditionally, when using deep neural networks for time series forecasting, we assume that a historical dataset $\mathcal{D}_{\text{his}} = \{\mathbf{x}_1, \mathbf{x}_2, \dots, \mathbf{x}_{T} \}$ is available, where the data spans from time 1 to time $T$, and each observation $\mathbf{x}_t \in \mathbb{R}^C$ for $t = 1, 2, \dots, T$, with $C$ representing the number of time series. In the conventional setup, we use $\mathcal{D}_{\text{his}}$ to train a neural network with parameters $\theta$, denoted as $f_\theta(\cdot)$, to perform the multivariate time series forecasting task.

In time series forecasting, our goal is to predict future values based on historical data. Specifically, given a historical dataset $\mathcal{D}_{\text{his}}$, we aim to train a neural network model $f_\theta(\cdot)$ to output predictions for the next $h$ time steps, i.e.,
\begin{equation}
   \{\hat{\mathbf{x}}_{t+1}, \hat{\mathbf{x}}_{t+2}, \cdots, \hat{\mathbf{x}}_{t+h}\} = f_\theta\left(\{\mathbf{x}_{t-l+1}, \mathbf{x}_{t-l+2}, \cdots, \mathbf{x}_{t}\}\right).
\end{equation}
We denote the predicted values $\{\hat{\mathbf{x}}_{t+1}, \hat{\mathbf{x}}_{t+2}, \cdots, \hat{\mathbf{x}}_{t+h}\}$ as $\hat{\mathbf{Y}}_t \in \mathbb{R}^{h \times C}$, and the historical values $\{\mathbf{x}_{t-l+1}, \mathbf{x}_{t-l+2}, \cdots, \mathbf{x}_{t}\}$ as ${\mathbf{X}}_t \in \mathbb{R}^{l \times C}$. Traditionally, the optimal model $f_{\theta^\ast}$ can be determined by minimizing the empirical risk through gradient descent on sampled data from the fixed training dataset $\mathcal{D}_{\text{his}}$. Specifically, we perform optimization by sampling samples from $\mathcal{D}_{\text{his}}$, and minimizing the following objective:
\begin{equation}
\theta^\ast = \text{argmin}_{\theta} \frac{1}{N_k} \sum_{i=1}^{N_k} \mathcal{L}\left(f_\theta(\mathbf{X}_i), \mathbf{Y}_i\right),
\label{eq:training}
\end{equation}
where $N_k$ is the number of samples drawn from the historical dataset $\mathcal{D}_{\text{his}}$, and $\mathcal{L}(\cdot)$ is the loss function, $\mathbf{Y}_t$ is the true observation of $\{{\mathbf{x}}_{t+1}, {\mathbf{x}}_{t+2}, \cdots, {\mathbf{x}}_{t+h}\}$

During the inference phase, assume that we have observed new data after time $T$. We will use the model $f_{\theta^\ast}$, obtained from the training dataset, to make predictions. This results in:
\begin{equation}
   \hat{\mathbf{Y}}_t = f_{\theta^\ast}\left(\mathbf{X}_{t}\right), \quad \forall t > T.
   \label{eq:inference}
\end{equation}

For a foundation time series model, the training dataset typically comes from multiple domains. It can be represented as $\mathcal{D}_\text{fdt} = \{ \mathcal{D}_1, \mathcal{D}_2, \dots, \mathcal{D}_N \}$, where $\mathcal{D}_n$ denotes the historical training dataset collected from the $n$-th domain. The foundation model is then typically tested on a target domain, $\mathcal{D}_\text{tgt}$, which is generally not part of $\mathcal{D}_\text{fdt}$, through zero-shot testing. Alternatively, few-shot testing can be performed, where the model is fine-tuned on $\mathcal{D}_\text{tgt}$ using Equation~\eqref{eq:training}, and then used for predictions on future observations from the target domain using Equation~\eqref{eq:inference}.

\subsection{Continual Learning Pipeline}

In this paper, we argue that the traditional pipeline described above is not suitable for practical time series forecasting models and foundation model applications, as it overlooks the fact that time series data is continuously collected over time. More importantly, various studies have shown that, over time, time series data is often accompanied by \textbf{temporal distribution shifts}. As the distribution of data evolves, a model trained on outdated, static datasets is no longer suitable for current prediction tasks. Formally, let the joint data distribution at time \(t\) be \(P_t(\mathbf{X}, \mathbf{Y})\), and assume that this distribution changes over time. A model \(f_{\theta^\ast}\) trained on an earlier dataset \(P_{\mathcal{D}_\text{his}}(\mathbf{X}, \mathbf{Y})\) will face degraded performance when applied to data from a new distribution \(P_t(\mathbf{X}, \mathbf{Y})\), as:
\begin{equation}
\mathbb{E}_{(\mathbf{X}, \mathbf{Y}) \sim P_{\mathcal{D}_\text{his}}} \left[ \mathcal{L}(f_{\theta^\ast}(\mathbf{X}), \mathbf{Y}) \right] \ll \mathbb{E}_{(\mathbf{X}, \mathbf{Y}) \sim P_t} \left[ \mathcal{L}(f_{\theta^\ast}(\mathbf{X}), \mathbf{Y}) \right].
\end{equation}

\begin{figure}
    \centering
    \includegraphics[width=1\linewidth]{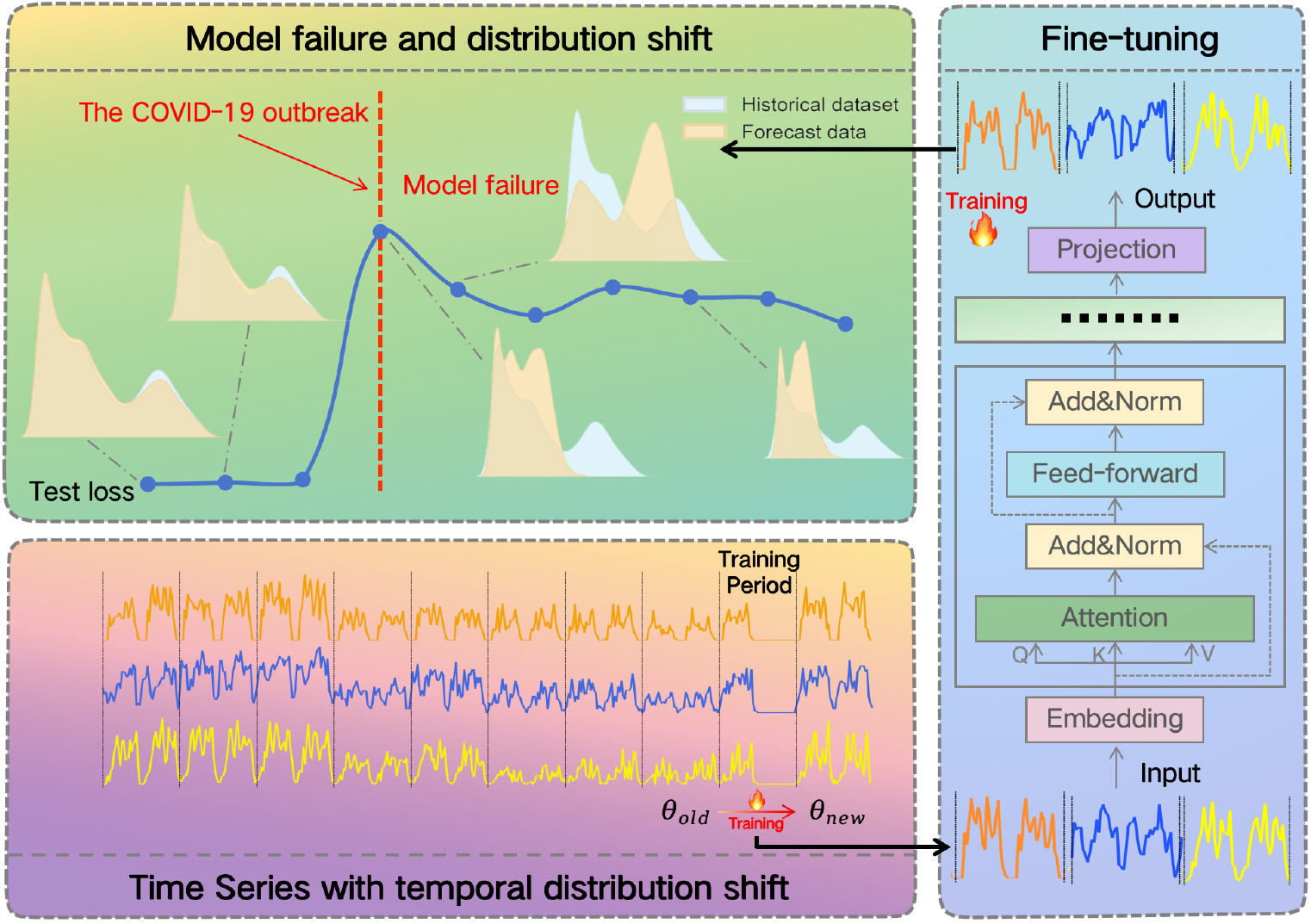}
    \caption{\textbf{Challenges in Time Series Modeling.} The figure illustrates model failure caused by temporal distribution shifts, such as the COVID-19 outbreak, and the divergence between historical and forecast data distributions.}
    \label{fig:model_failure}
\end{figure}

This raises a critical issue that warrants in-depth investigation: \textbf{will deep time series forecasting models eventually fail over time, particularly when time series data experiences significant temporal distribution shifts due to external factors or other causes?} We provide an example in Figure~\ref{fig:model_failure}, where we periodically fine-tune a time series using a Transformer with newly collected data. However, after the outbreak of COVID-19, a significant distribution shift occurred between the new and old data. When we continued fine-tuning the model with the post-outbreak data, the model's test error began to increase significantly.

Given the existence of numerous models based on various architectures, such as MLP, Transformer, TCN, and GNN, \textbf{is this phenomenon inherent to all deep learning models?}  More importantly: \textbf{Can foundation models avoid this issue?} If foundation models also become ineffective after a certain period, the consequences could be catastrophic, as their retraining would demand significant computational resources.

One might consider periodically retraining the model using the $p$-th newly collected dataset $\mathcal{D}_{p} = \{\mathbf{x}_{T+pS}, \mathbf{x}_{T+pS+1}, \cdots,  \mathbf{x}_{T+pS+p-1}\}$, with $S$ being the retraining time span, and fine-tuning the old model parameters $\theta_{\text{old}}$ as follows:
\begin{equation}
\theta_{\text{new}}^\ast = \text{argmin}_{\theta} \frac{1}{|\mathcal{D}_{p}|} \sum_{(\mathbf{X}_i, \mathbf{Y}_i) \in \mathcal{D}_{p}} \mathcal{L}(f_\theta(\mathbf{X}_i), \mathbf{Y}_i),
\label{eq:continual_training}
\end{equation}
where $\mathcal{D}_{p}$ represents the new dataset collected during the time span from $T+pS$ to $T+pS+p-1$, and $\mathcal{L}(\cdot)$ is the loss function used for training, $\theta_{\text{old}} = \theta^{\ast}$ when $p = 0$. If such a training method is feasible, we would only need to fine-tune a foundation model with a small amount of newly collected data, enabling it to adapt to the new distribution. Unfortunately, it is reported that the training method in Equation~\eqref{eq:continual_training} not only suffers from catastrophic forgetting but also leads to a loss of plasticity\cite{dohare2024loss}. 

Catastrophic forgetting occurs when the model forgets previously learned information while adapting to new data. This can be mathematically expressed as:
\begin{equation}
    \mathbb{E}_{(\mathbf{X}, \mathbf{Y}) \sim P_{\text{old}}} \left[ \mathcal{L}(f_{\theta_{\text{new}}}(\mathbf{X}), \mathbf{Y}) \right] > \mathbb{E}_{(\mathbf{X}, \mathbf{Y}) \sim P_{\text{new}}} \left[ \mathcal{L}(f_{\theta_{\text{new}}}(\mathbf{X}), \mathbf{Y}) \right],
\end{equation}
where $P_{\text{old}}$ and $P_{\text{new}}$ represent the old and new data distributions, respectively. Catastrophic forgetting happens when the model's performance on the old data distribution deteriorates significantly after retraining on the new data. One might think that this isn’t a big issue for time series models, as one could assume that data points close in time likely have similar distributions. However, time series are quite complex. For example, large economic cycles can last for decades, meaning that past distributions could reappear at some point in the future. Therefore, the model fine-tuned on the newly collected training data may not necessarily have a performance advantage for a period closer to the future, due to the phenomenon of catastrophic forgetting.

Loss of plasticity refers to the model’s inability to effectively adapt to new tasks or data distributions after a certain amount of training. This can be described mathematically as:
\begin{equation}
    \lim_{p \to \infty} \mathbb{E}_{(\mathbf{X}, \mathbf{Y}) \sim P_{\text{new}}} \left[ \mathcal{L}(f_{\theta_{\text{new}}}(\mathbf{X}), \mathbf{Y}) \right] \to \mathcal{L}(f_{\theta_{\text{old}}}(\mathbf{X}), \mathbf{Y}),
\end{equation}
where, after repeated retraining with new data, the model's ability to improve its performance on the new distribution diminishes, indicating a loss of plasticity. This phenomenon is even more discouraging. Suppose we invest enormous computational resources to train a foundation time series model using massive datasets from multiple domains. We may want to fine-tune it continuously with minimal resources, allowing it to adapt to the changes brought about by time. However, the loss of plasticity tells us that at some point, this model will lose its ability to continue adapting to the new time series distribution. This means that periodically, we might need to completely retrain the foundation model.

So, in this paper, we investigate another important issue: \textbf{Can traditional deep learning models and foundation models avoid the problems of catastrophic forgetting and loss of plasticity?}




\section{Evaluation Pipeline}

For Time Series Foundation Models, during the application process, we often work with pre-existing pre-trained parameters \( \theta_{\text{fdt}} \). These parameters are obtained through large-scale pre-trained datasets and considerable computational power, endowing the model with strong zero-shot learning capability. Therefore, for any input \( \mathbf{X} \), we can evaluate the performance of the foundation model in a zero-shot setting. In time series forecasting, it is common practice to assess model error by transforming the output data distribution to \( \mathcal{N}(0, 1) \) and then calculating the model's error, often using metrics such as Mean Squared Error (MSE).

When we use MSE to measure the difference between two distributions, particularly one being the standard Gaussian distribution \( \mathcal{N}(0, 1) \), we are essentially quantifying the deviation between the distributions' values. We have
\begin{equation}
   \text{MSE}(\hat{\mathbf{Y}}, \mathbf{Y}) = \mathbb{E}[(\hat{\mathbf{Y}} - \mathbf{Y})^2],
\end{equation}
where $\hat{\mathbf{Y}}$ is the predicted value, and $\mathbf{Y}$ is the true value obeying \( \mathcal{N}(0, 1) \). Assume \( \hat{\mathbf{Y}} \) comes from another distribution (e.g., \( \hat{\mathbf{Y}} \sim \mathcal{N}(\mu, \sigma^2) \) and $\hat{\mathbf{Y}}$ and $\mathbf{Y}$ are mutually independent. Then, MSE can be expanded as a measure of:
\begin{equation}
    \mathbb{E}[(\hat{\mathbf{Y}} - \mathbf{Y})^2] = \mathbb{E}[\mathbf{Y}^2] - 2\mathbb{E}[\hat{\mathbf{Y}} \mathbf{Y}] + \mathbb{E}[\hat{\mathbf{Y}}^2].
\end{equation}

For the standard Gaussian distribution \( \mathbf{Y} \sim \mathcal{N}(0, 1) \), we know that \( \mathbb{E}[\mathbf{Y}^2] = 1 \) and \( \mathbb{E}[\mathbf{Y}] = 0 \). Thus, the MSE becomes a measure of:
\begin{equation}
   \text{MSE}(\hat{\mathbf{Y}}, \mathbf{Y}) \approx 1 - 0 + (\sigma^2 + \mu^2) = 1 + \sigma^2 + \mu^2. 
\end{equation}

Through simple analysis, we can see that if we calculate the MSE error for the normalized true values \( \mathbf{Y} \), we are essentially measuring the error between the model's output distribution mean and 0, along with the square of the output variance, which is a measure of uncertainty. Therefore, assuming a zero-shot MSE error, we have:
\begin{equation}
    \text{MSE}^{\text{zero}} \approx 1 + \sigma_{\text{fdt}}^2 + \mu_{\text{fdt}}^2,
\end{equation}
where \( \sigma_{\text{fdt}}^2 \) and \( \mu_{\text{fdt}}^2 \) represent the variance and mean of the foundation model's output distribution, respectively.

Now, assume we have collected new data and apply the incremental fine-tuning in Algorithm~\ref{alg:fine_tuning}, dividing the dataset into \( p = 0, 1, \dots, P-1 \) parts based on time. Then, for each of the \( P \) test sets, we normalize the output data and test the MSE error. We will obtain \( P \) measures of \( 1 + \sigma_{\text{p}}^2 + \mu_{\text{p}}^2 \), with \( p \) indicating the index of the fine-tuning datasets. We can then obtain the first measure:
\begin{equation}
   \text{R}^{\text{zero}}_p = \frac{\text{MSE}^{\text{inc}}_{\text{p}}}{\text{MSE}^{\text{zero}}_{\text{p}}}, \quad \forall p = 0,1,\cdots, P-1.
\end{equation}
where \( \text{MSE}^{\text{inc}}_{\text{p}} \) is the MSE for the \( p \)-th incremental fine-tuning dataset, and \( \text{MSE}^{\text{zero}}_{\text{p}} \) is the MSE obtained in the zero-shot setting for the \( p \)-th dataset. Thus, the ratio \( \text{R}^{\text{zero}}_p \) captures the relative change in the error between the fine-tuned model and the zero-shot model in terms of their means and variances. For the foundation model, we hope that the model can consistently maintain \( \text{R}^{\text{zero}}_p < 1 \). If the model can achieve a decreasing trend in \( \text{R}^{\text{zero}}_p \) as \( p \) increases (in the long term), this indicates that the model, in a certain sense, has the ability to continuously optimize in a specific domain, without being affected by loss of plasticity and catastrophic forgetting.

\begin{algorithm}
\caption{Incremental Fine-Tuning Algorithm for Deep Time Series Model}
\label{alg:fine_tuning}
\begin{algorithmic}[1]
\STATE Load an old model $f_{\theta^\text{old}}$ fine-tuned by $\mathcal{D}_{p-1}$.
\STATE Given a new dataset $\mathcal{D}_{p} = \{\mathbf{x}_{T+pS}, \mathbf{x}_{T+pS+1}, \cdots,  \mathbf{x}_{T+pS+p-1}\}$.
\FOR{each batch $\mathcal{B} \subseteq \mathcal{D}_{\text{p}}$}
    \STATE Compute model output $\hat{\mathbf{y}} = f_\theta(\mathcal{B})$
    \STATE Compute MSE loss $\mathcal{L}_{\text{mse}} = \frac{1}{|\mathcal{B}|} \sum_{i} \|\hat{\mathbf{y}}_i - \mathbf{y}_i\|^2$
    \STATE Update model parameters $\theta \gets \theta - \eta \nabla_\theta \mathcal{L}_{\text{mse}}$
\ENDFOR
\RETURN Fine-tuned model $f_\theta$
\end{algorithmic}
\end{algorithm}

One may think that the method used in Algorithm~\ref{alg:fine_tuning} is merely an incremental fine-tuning approach, which might not fully unleash the potential of the foundation model on new data. Therefore, we adopted a full fine-tuning approach to simulate the model's strongest capability. In Algorithm~\ref{alg:full_training}, after a certain period of time, we train a completely new model. This approach, when applied to smaller models, yields the optimal function approximator for the test dataset. However, it is important to note that, for the foundation model, since it has already been trained on large-scale pre-trained datasets, there is a possibility that full training might lead to a decrease in performance. Moreover, this approach is also not ideal for us, because fine-tuning the foundation model is computationally expensive. Re-training with the entire collected dataset each time is very unfriendly for users with limited computational resources. Assume that the MSE for the \( p \)-th full training is \( \text{MSE}^{\text{full}}_{\text{p}} \). Then, for both the foundation model and smaller models, we have an additional measure:
\begin{equation}
   \text{R}^{\text{full}}_p = \frac{\text{MSE}^{\text{inc}}_{\text{p}}}{\text{MSE}^{\text{full}}_{\text{p}}}, \quad \forall p = 0, 1, \dots, P-1.
\end{equation}
This reflects the ratio of the error between the model's incremental fine-tuning and the most optimal model (only for small model). For \( \text{R}^{\text{full}}_p \), we hope that it stays close to 1. Of course, achieving a value less than 1 is challenging, but if its long-term trend is stable around 1 or decreasing, it would also indicate that the model has strong continual learning capability.

\begin{algorithm}
\caption{Full Training Algorithm for Deep Time Series Model}
\label{alg:full_training}
\begin{algorithmic}[1]
\STATE Given the historical dataset $\mathcal{D}_{\text{his}} = \{\mathcal{D}_{0}, \mathcal{D}_{1}, \cdots, \mathcal{D}_{p-1}\}$ and the newly collected dataset $\mathcal{D}_{p} = \{\mathbf{x}_{T+pS}, \mathbf{x}_{T+pS+1}, \cdots,  \mathbf{x}_{T+pS+p-1}\}$.
\STATE Combine datasets to form the complete dataset: $\mathcal{D}_{\text{all}} = \mathcal{D}_{\text{his}} \cup \mathcal{D}_{\text{p}}$.
    \IF{The model is foundational (e.g., Time-MoE, Chronos)}
        \STATE Load a pre-trained model $f_\theta$ with parameters $\theta_{\text{pretrained}}$
    \ELSE
        \STATE Initialize a new small model $f_\theta$ with random weights.
    \ENDIF
\FOR{each batch $\mathcal{B} \subseteq \mathcal{D}_{\text{all}}$}
    \STATE Compute model output $\hat{\mathbf{y}} = f_\theta(\mathcal{B})$.
    \STATE Compute MSE loss: $\mathcal{L}_{\text{mse}} = \frac{1}{|\mathcal{B}|} \sum_{i} \|\hat{\mathbf{y}}_i - \mathbf{y}_i\|^2$.
    \STATE Update model parameters: $\theta \gets \theta - \eta \nabla_\theta \mathcal{L}_{\text{mse}}$.
\ENDFOR
\RETURN Trained model $f_\theta$
\end{algorithmic}
\end{algorithm}

For the foundation model, we can also compare \( \text{MSE}^{\text{full}}_{\text{p}} \) and \( \text{MSE}^{\text{zero}}_{\text{p}} \), which allows us to assess the model's performance after full training versus its performance in the zero-shot setting. We have
\begin{equation}
   \text{R}^{\text{fz}}_p = \frac{\text{MSE}^{\text{full}}_{\text{p}}}{\text{MSE}^{\text{zero}}_{\text{p}}}, \quad \forall p = 0, 1, \dots, P-1.
\end{equation}
Similarly, we hope that \( \text{R}^{\text{fz}}_p \) remains consistently below 1, with a long-term decreasing trend. This would indicate that as the model encounters more samples from unseen domains, its adaptability improves.




\section{Experiments}

\begin{figure}
    \centering
    \includegraphics[width=1\linewidth]{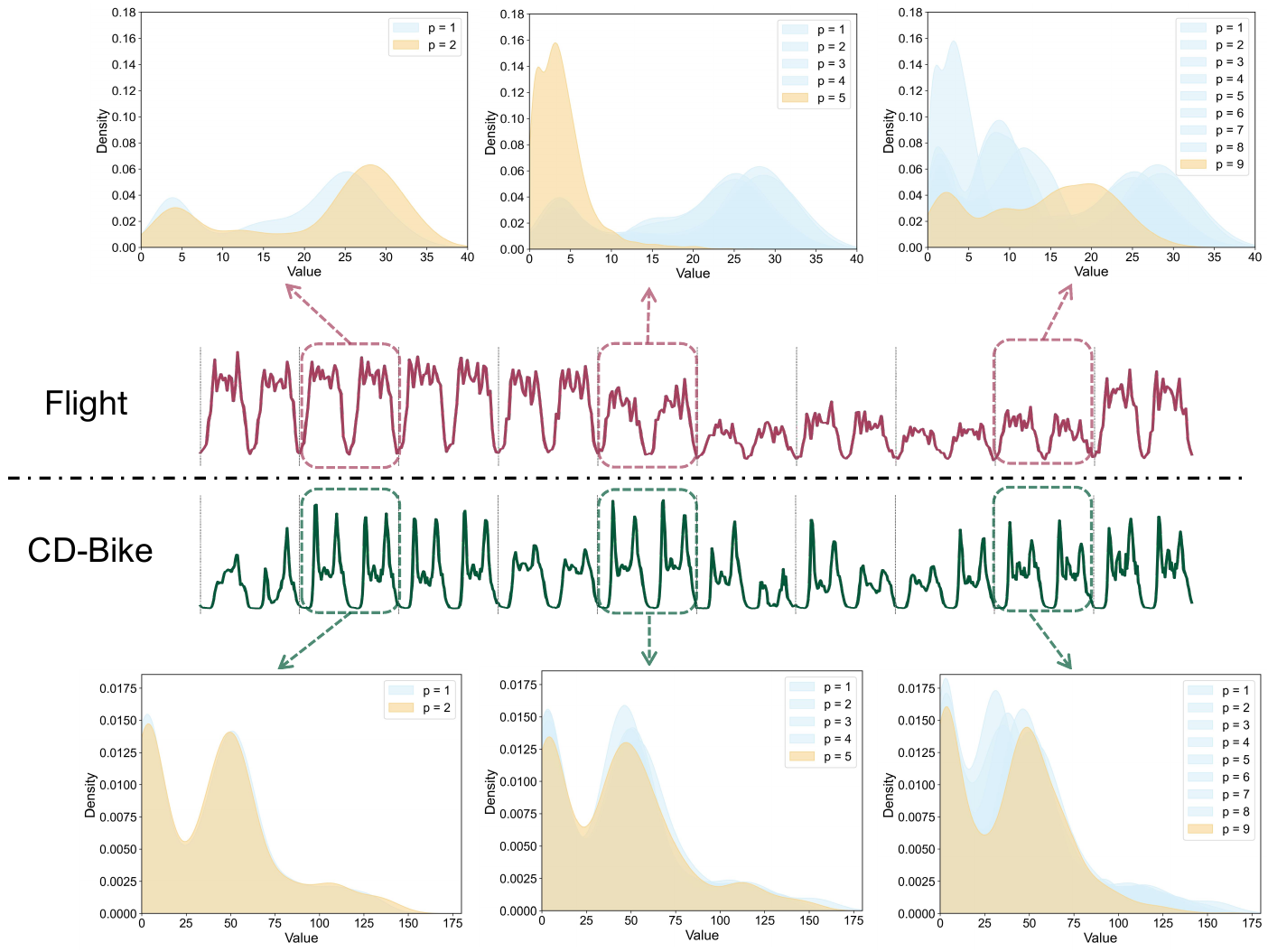}
    \caption{{Data distribution shifts in the Flight and CD-Bike datasets.} }
    \label{fig:data distribution}
\end{figure}

\textbf{Datasets. } To avoid data contamination from foundation models' training, we carefully selected two datasets: a proprietary Chengdu bike-sharing dataset and a lesser-known dataset that we verified were not included in large language models' training datasets. These datasets are: \textbf{Flight}~\cite{cai2024msgnet}, characterized by major domain distribution shifts, and \textbf{CD-Bike}~\cite{cheng2024rethinking}, exhibiting minor shifts. Each dataset was chronologically partitioned into \(P = 10\) subsets to simulate real-world continual learning scenarios. The distribution characteristics are shown in Figure~\ref{fig:data distribution}. Obviously, the Flight dataset shows significant shifts (for instance, Partitions 5 and 9 exhibit substantial distribution shifts due to COVID-19 outbreak), while the CD-Bike dataset remains relatively stable over time.

\begin{table}[ht]
\centering
\caption{Statistics of Datasets}
\begin{tabular}{lcccc}
\hline
\textbf{Datasets}  & \textbf{Variables} & \textbf{Time Range} & \textbf{Interval} & \textbf{Timesteps} \\ \hline
Flight            & 7               & 2019/1/1--2021/12/31      & 1h        & 26304               \\
CD-Bike           & 34               & 2023/3/13--2023/11/3      & 0.5h      & 11280              \\ \hline
\end{tabular}
\label{tab:dataset_stats}
\end{table}


\textbf{Baselines. }
We compared the performance of small models and foundation models across the three training strategies. Small models (\textbf{DLinear}~\cite{zeng2023transformers}, \textbf{PatchTST}~\cite{nie2022time}, \textbf{iTransformer}~\cite{liu2023itransformer}) were evaluated using Incremental Fine-Tuning and Full Training methods, while foundation models (\textbf{Time-MoE}~\cite{shi2024time}, \textbf{Chronos}~\cite{ansari2024chronos}) were additionally tested using Zero-Shot evaluation. All models followed the same data preprocessing pipeline to ensure fair comparison. 

\textbf{Setups. }
We divided \textbf{Flight} and \textbf{CD-Bike} into \(P = 10\) subsets to simulate continual learning scenarios. For each subset \(p\), the data was partitioned into training, validation, and test sets in a \(6:2:2\) ratio. All models were trained with an input context length of 96 and a prediction length of 96, except for \textbf{Chronos}, which has a maximum prediction length of 64 due to its architectural constraints. The optimizer used for all models was AdamW with a learning rate of \(1 \times 10^{-4}\), and each model was trained for 10 epochs. The performance of all models was evaluated using the Mean Squared Error (MSE) metric. All experiments were conducted on an NVIDIA GeForce RTX 3090 GPU to ensure efficient training and evaluation.

\begin{figure}
    \centering
    \includegraphics[width=0.75\linewidth]{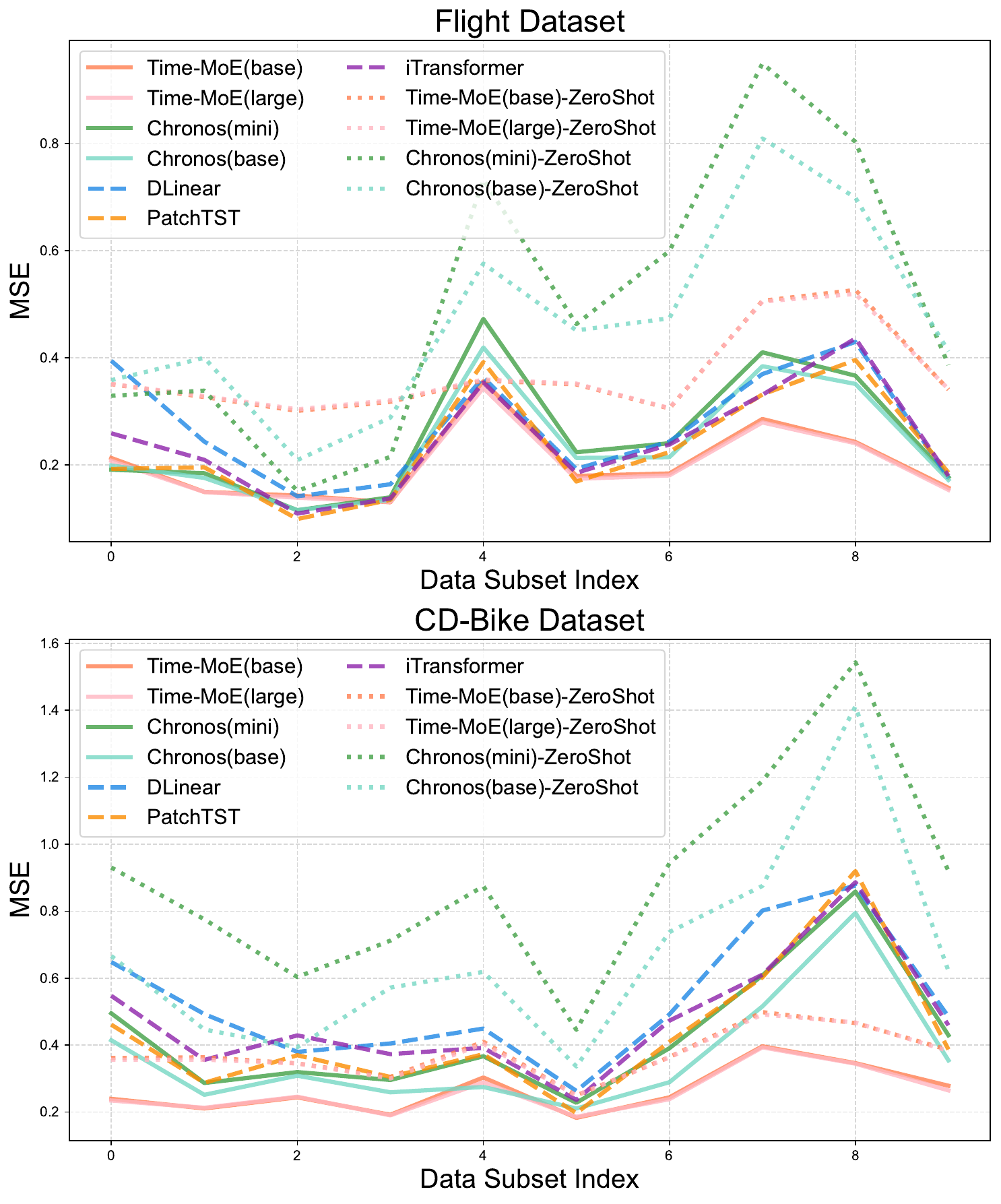}
    \caption{\textbf{Incremental Fine-Tuning Algorithm results for the Flight (top) and CD-Bike (bottom) datasets.} Solid lines represent foundation models' fine-tuning performance, dashed lines show small models' fine-tuning results, and dotted lines indicate foundation models' zero-shot performance.}
    \label{fig:Fine-Tuning}
\end{figure}

\subsection{Performance Comparison }

\textbf{Incremental Fine-Tuning Results.} The performance comparison of models using Incremental Fine-Tuning and foundation models in zero-shot settings is shown in Figure~\ref{fig:Fine-Tuning}. On the \textbf{Flight} dataset with significant distribution shifts, Time-MoE achieved superior performance, followed by small models, while Chronos showed the lowest performance under fine-tuning. However, fine-tuned foundation models demonstrated substantial improvement over their zero-shot performance, highlighting the effectiveness of adaptation. On the \textbf{CD-Bike} dataset with minor distribution shifts, Time-MoE maintained its leading performance, with Chronos ranking second. Notably, Time-MoE's zero-shot performance surpassed both fine-tuned small models and Chronos, demonstrating its robust generalization capabilities. Time-MoE's superior performance over Chronos can be attributed to its specialized MoE architecture and larger parameter size.

\begin{figure}
    \centering
    \includegraphics[width=0.75\linewidth]{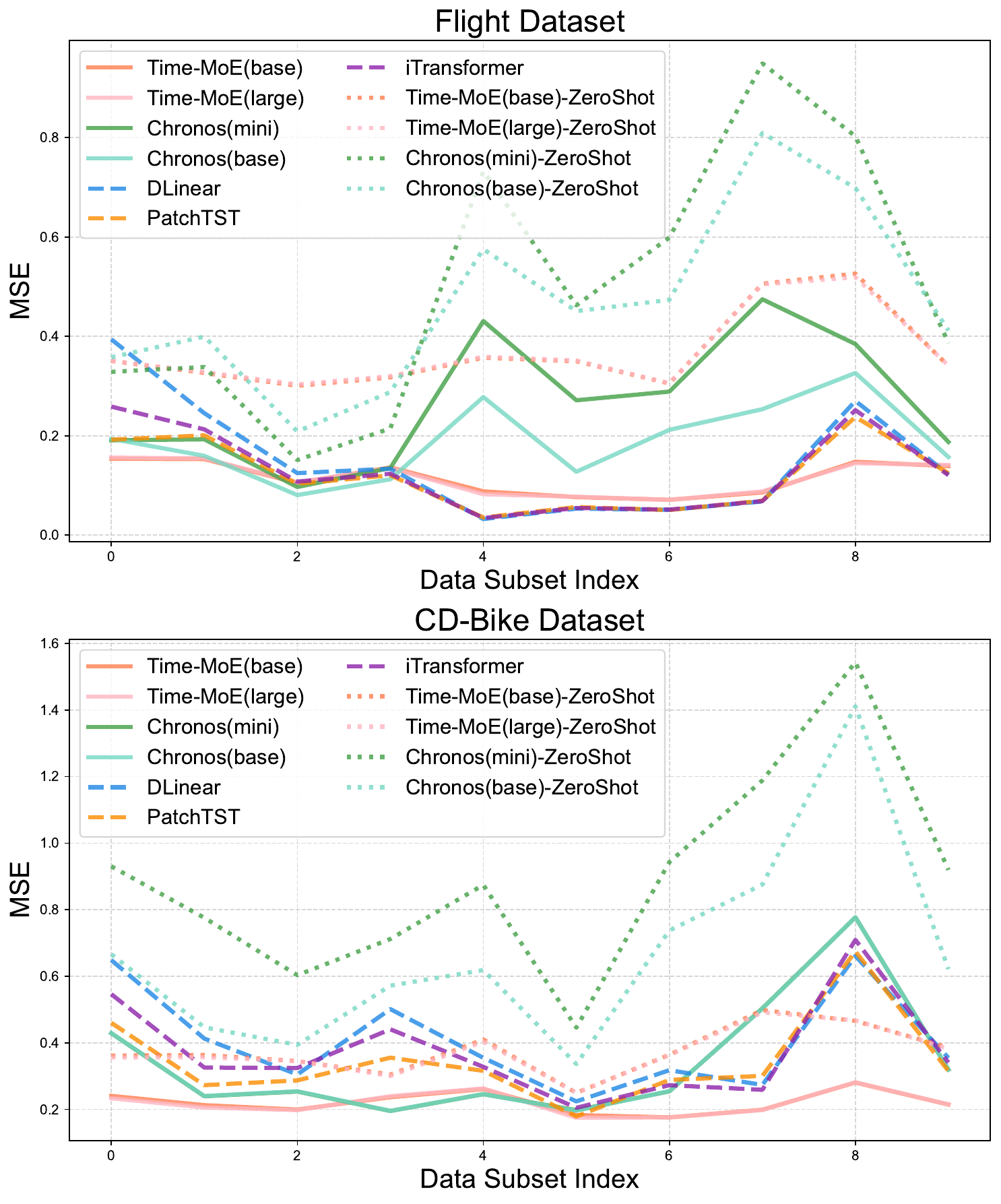}
    \caption{\textbf{Full Training Algorithm results for the Flight (top) and CD-Bike (bottom) datasets.} Solid lines represent foundation models' full training performance, dashed lines show small models' performance, and dotted lines indicate foundation models' zero-shot results.}
    \label{fig:full}
\end{figure}

\textbf{Full Training Results.} The Full Training results are shown in Figure~\ref{fig:full}. On the \textbf{Flight} dataset, smaller models occasionally outperform larger architectures, with Time-MoE consistently surpassing Chronos. Full Training substantially reduces MSE compared to zero-shot performance, demonstrating the benefits of comprehensive training. For the \textbf{CD-Bike} dataset, Time-MoE maintains superior performance, followed by Chronos, while smaller models show inferior results. Notably, Time-MoE's zero-shot performance sometimes exceeds the Full Training results of smaller models. Small models demonstrate significantly better performance with Full Training compared to fine-tuning, indicating that incremental approaches fail to fully utilize their capacity.

These results indicate that fine-tuning or full-training state-of-the-art foundation models on new datasets consistently outperforms smaller models. This suggests that leveraging and adapting industry-scale pre-trained models may be more valuable than training specialized small models for the target time-series.

\subsection{Continuous learning ability}

\begin{figure}
    \centering
    \includegraphics[width=0.75\linewidth]{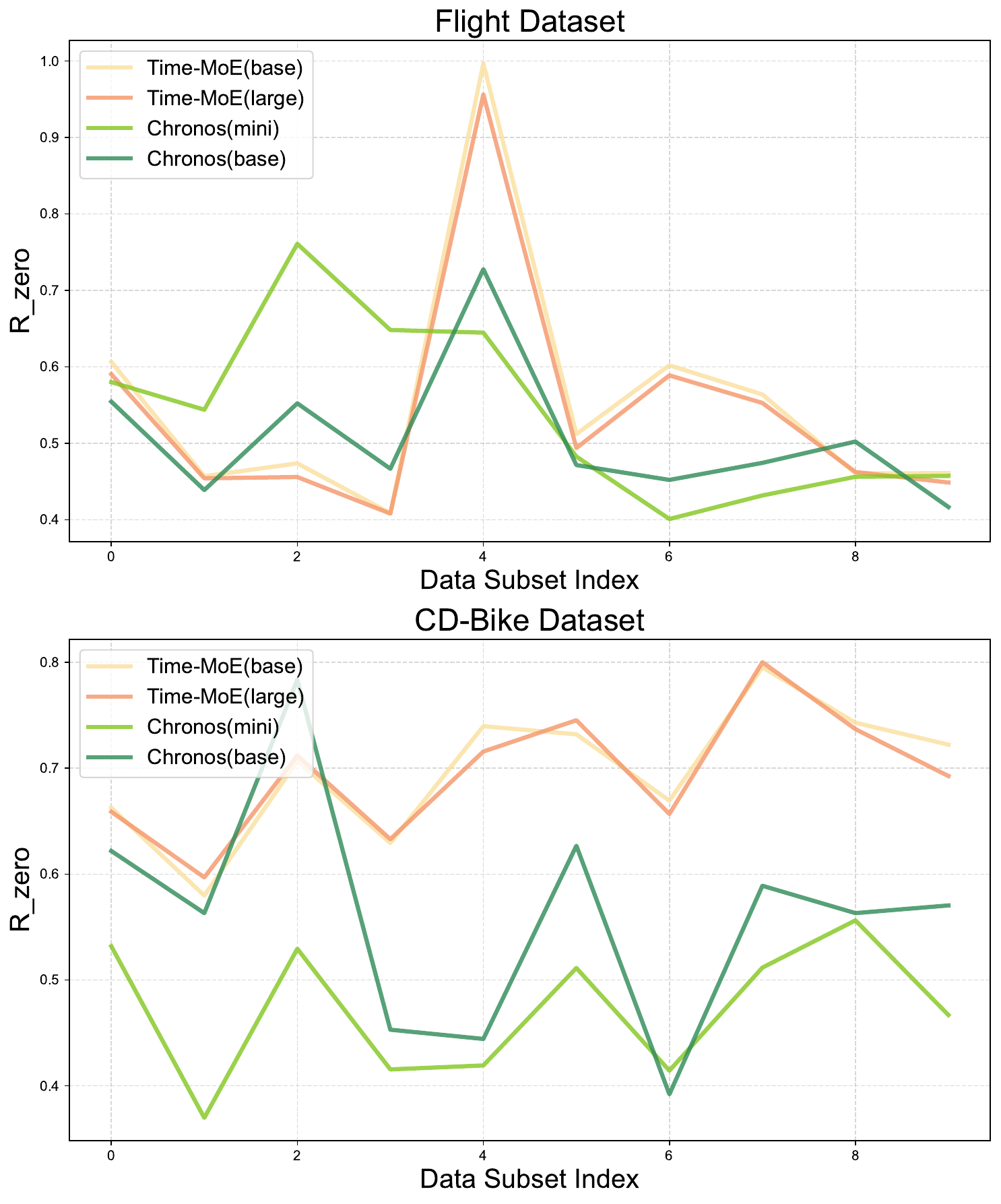}
    \caption{$R^{\text{zero}}_p$ for the Flight (top) and CD-Bike (bottom) datasets.}
    \label{fig:R-zero}
\end{figure}

\textbf{Metric $R^{\text{zero}}_p$: Incremental Fine-Tuning vs. Zero-Shot Performance.} The $R^{\text{zero}}_p$ values, shown in Figure~\ref{fig:R-zero}, highlight the improvement of incremental fine-tuning over zero-shot performance, reflecting models' ability to optimize through continual learning. On the \textbf{Flight} dataset, both Time-MoE and Chronos exhibit R-zero values below 0.7 across most subsets, indicating their capacity for continual optimization of the model. Both Time-MoE and Chronos exhibit a noticeable spike at \(p = 4\), which is likely due to the onset of data distribution shifts at this point caused by COVID-19 outbreak. Since the models have not yet fully adapted to the new distribution during Incremental Fine-Tuning, their predictive performance suffers, resulting in suboptimal fine-tuning outcomes for foundation models. On the \textbf{CD-Bike} dataset, Chronos (mini) achieves the lowest $R^{\text{zero}}_p$ value. Chronos generally exhibits lower $R^{\text{zero}}_p$ values compared to Time-MoE, indicating that Incremental Fine-Tuning significantly enhances Chronos's adaptability. This suggests that Chronos demonstrates stronger continual learning capabilities in this dataset with relatively stable distribution. Moreover, we do not observe a consistent decrease in $R^{\text{zero}}_p$ with increasing $p$, indicating that our raw incremental learning approach does not progressively enhance foundation models' domain adaptation capabilities.

\begin{figure}
    \centering
    \includegraphics[width=0.75\linewidth]{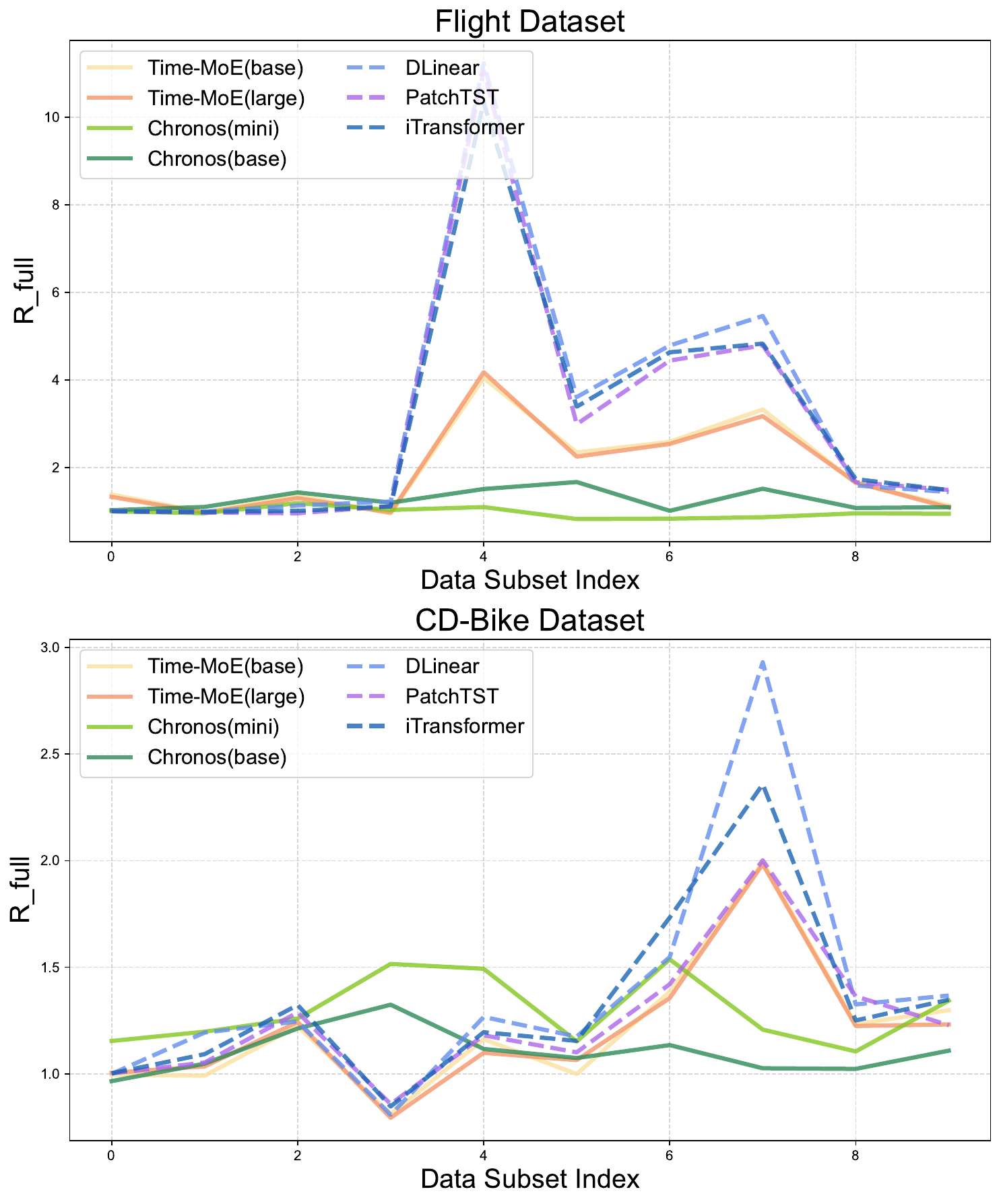}
    \caption{$R^{\text{full}}_p$ for the Flight (top) and CD-Bike (bottom) datasets.}
    \label{fig:R-full}
\end{figure}

\textbf{Metric $R^{\text{full}}_p$: Incremental Fine-Tuning vs. Full Training.} As illustrated in Figure~\ref{fig:R-full}, $R^{\text{full}}_p$ evaluates how closely incremental fine-tuning performance approaches that of full training, reflecting models' progressive learning capacity and temporal plasticity. On both Flight and CD-Bike datasets, Chronos and Time-MoE demonstrate significantly lower $R^{\text{full}}_p$ values compared to small models, indicating foundation models can achieve near full-training performance through continuous learning on limited distribution-shifted data. In contrast, small models lack this capability, exhibiting plasticity loss in temporal continual learning scenarios. This highlights the superior temporal plasticity of foundation models.

On the Flight dataset, both foundation and small models show a spike in $R^{\text{full}}_p$ at \(p = 4\). This spike is likely due to the onset of a distribution shift, where incremental fine-tuning struggles to fully capture the new data characteristics, while full training effectively adapts to the shift. In contrast, the CD-Bike dataset, with its smaller distribution shifts, exhibits smoother and consistently lower $R^{\text{full}}_p$ values (below 3). Notably, small models on the Flight dataset experience a sharp increase at \(p = 4\), with $R^{\text{full}}_p$ reaching as high as 10, reflecting their difficulty in handling substantial distribution shifts during incremental fine-tuning. Moreover, their inferior $R^{\text{full}}_p$ compared to foundation models emerges after the 4-th dataset, suggesting their loss of plasticity likely stems from abrupt distribution shifts in the time series. This phenomenon demands attention: while incremental learning struggles to fully adapt models to distribution-shifted temporal data, full retraining with the combined dataset yields superior performance on new distributions. This gap highlights a critical future research direction: \textbf{developing effective fine-tuning strategies for time series models under significant distribution shifts}.

\begin{figure}
    \centering
    \includegraphics[width=0.75\linewidth]{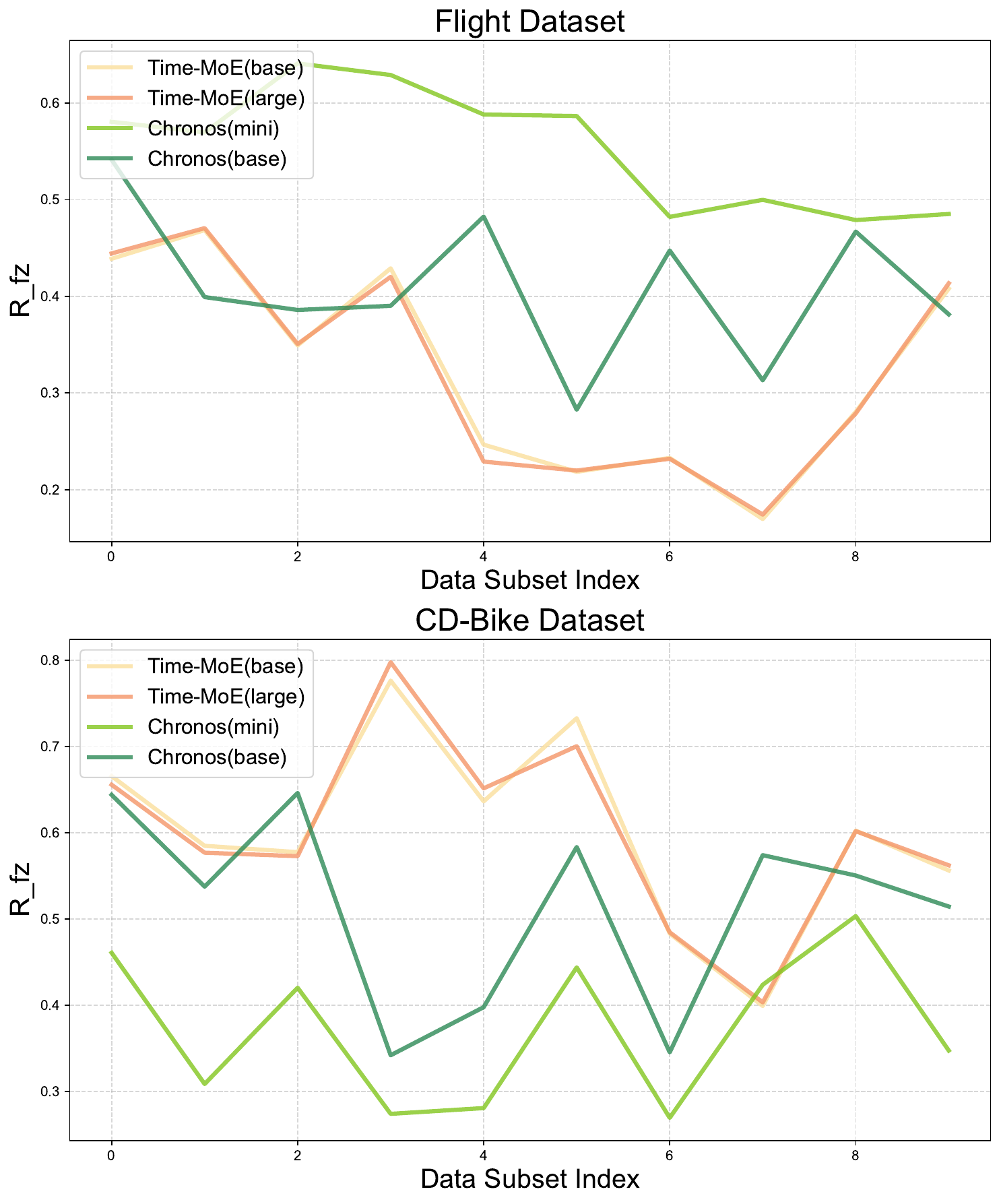}
    \caption{$R^{\text{fz}}_p$ for the Flight (top) and CD-Bike (bottom) datasets.}
    \label{fig:R-fz}
\end{figure}

\textbf{Metric $R^{\text{fz}}_p$: Full Training vs. Zero-Shot Performance.} This metric, shown in Figure~\ref{fig:R-fz}, reflects additional domain-specific data make the model more adaptable to the domain-specific task. On the Flight dataset, Time-MoE  achieves R-fz values of 0.2–0.4, outperforming Chronos models, which hover around 0.3–0.6. The foundation models on the Flight dataset no longer exhibit a spike, indicating that full training effectively captures the dataset's distribution shift and better adapts to the current prediction task. On the CD-Bike dataset, Chronos outperforms Time-MoE, suggesting that Chronos may be better suited for datasets with relatively minor distribution shifts. Moreover, Time-MoE, with its specialized MoE architecture, appears particularly well-suited for time series forecasting tasks involving significant data distribution shifts.

In summary, foundation models like Time-MoE and Chronos demonstrate robust continual learning capabilities and effectively mitigate plasticity loss, showing significant improvements through both Incremental Fine-Tuning and Full Training. In contrast, smaller models struggle to capture data characteristics through Incremental Fine-Tuning. The success of foundation models trained on large-scale data with larger architectures provides a promising foundation for addressing catastrophic forgetting and plasticity loss in continual learning and life-long learning scenarios. Interestingly, neither Time-MoE nor Chronos show significant performance differences across their mini, base, and large variants. This suggests that model size may have less impact on fine-tuning adaptability than anticipated, indicating that models achieve strong generalization capabilities once they reach a sufficient scale for training with large pre-training datasets.

\section{Conclusion}
This work presents the first comprehensive investigation of temporal plasticity in time series foundation models. Through extensive experiments on real-world datasets with varying distribution shifts, we demonstrate that foundation models like Time-MoE and Chronos exhibit superior continual learning capabilities compared to smaller models. Our key findings reveal that: (1) Foundation models effectively mitigate plasticity loss and catastrophic forgetting during incremental fine-tuning; (2) Model size beyond a certain threshold has diminishing returns on adaptability, suggesting pre-training data scale may be more crucial than architecture size; and (3) Raw incremental learning approaches still struggle to fully adapt to significant distribution shifts, highlighting the need for more robust fine-tuning strategies. These insights suggest that future research should focus on developing efficient adaptation mechanisms for foundation models rather than designing specialized small models. Our evaluation framework and metrics provide valuable tools for assessing and improving the temporal plasticity of time series models in practical applications.

\bibliographystyle{IEEEtran}

\end{document}